\documentclass[letterpaper, 10 pt, conference]{ieeeconf}  

\IEEEoverridecommandlockouts                              

\usepackage{epsfig} 
\usepackage{mathptmx} 
\usepackage{times} 
\usepackage{amsmath} 
\usepackage{amssymb}  
\usepackage[T1, OT1]{fontenc}
\DeclareTextSymbolDefault{\DH}{T1}
\usepackage{hyperref}
\usepackage{nicefrac}       
\usepackage[dvipsnames]{xcolor}
\usepackage{microtype}      
\usepackage{graphicx}
\usepackage{caption}
\usepackage{subcaption}
\usepackage{algpseudocode}
\usepackage{xspace}
\usepackage{wrapfig}
\usepackage{ctable}
\usepackage{multirow}
\usepackage{varwidth}
\usepackage{adjustbox}
\usepackage{arydshln}
\usepackage{physics}
\usepackage{hhline}
\usepackage{makecell}
\usepackage{booktabs}

\newcommand{\ours}{PnF}
\newcommand{\ourslong}{Plug-and-Forecast}
\newcommand\mypara[1]{\vspace{1.mm}\noindent\textbf{#1}}
\newcommand{\datasetfullname}{Waymo Open Motion Dataset}
\newcommand{\datasetabbr}{WOMD}
\newcommand{\colorhighlight}[1]{\textcolor{black}{#1}}  

\definecolor{custombox}{HTML}{ffcd55}
\definecolor{custombox2}{HTML}{00e89d}
\definecolor{vsaoutcolor}{HTML}{f4af91} 
\definecolor{scoutcolor}{HTML}{0078ff}
\definecolor{cvprblue}{rgb}{0.21,0.49,0.74}

\title{\LARGE \bf
Enhanced Motion Forecasting with Plug-and-Play\\Multimodal Large Language Models}

\author{Katie Luo$^\dag$ \quad Jingwei Ji$^{*\ddag}$%
\quad Tong He$^\ddag$ \quad Runsheng Xu$^\ddag$ \quad Yichen Xie$^\S$\\Dragomir Anguelov$^\ddag$ \quad Mingxing Tan$^\ddag$%
\thanks{* Corresponding author.}%
\thanks{$\dag$ Computer and Information Sciences Department, Cornell University, 
        {\tt\small \{kzl6\}@cornell.edu}. Work was done at Waymo.}%
\thanks{$\ddag$ Waymo LLC, 
        {\tt\small \{jingweij, simpleig, runshengxu, dragomir, tanmingxing\}@waymo.com}}%
\thanks{$\S$ UC Berkeley, 
        {\tt\small \{yichen\_xie\}@berkeley.edu}. Work was done at Waymo.}%
}

\begin{document}

\maketitle
\thispagestyle{empty}
\pagestyle{empty}



\begin{abstract}
Current autonomous driving systems rely on specialized models for perceiving and predicting motion, which demonstrate reliable performance in standard conditions. However, generalizing cost-effectively to diverse real-world scenarios remains a significant challenge. To address this, we propose \ourslong{} (\ours{}), a plug-and-play approach that augments existing motion forecasting models with multimodal large language models (MLLMs). \ours{} builds on the insight that natural language provides a more effective way to describe and handle complex scenarios, enabling quick adaptation to targeted behaviors. We design prompts to extract structured scene understanding from MLLMs and distill this information into learnable embeddings to augment existing behavior prediction models. 
Our method leverages the zero-shot reasoning capabilities of MLLMs to achieve significant improvements in motion prediction performance, while requiring no fine-tuning—making it practical to adopt.
We validate our approach on two state-of-the-art motion forecasting models using the \datasetfullname{} and the nuScenes Dataset, demonstrating consistent performance improvements across both benchmarks.
\end{abstract}    
\section{Introduction}
\label{sec:intro}





Since the first DARPA challenge, autonomous driving has progressed significantly, evolving into robo-taxis and driver assistance systems for mass consumer use \cite{waymo_2024,tesla_autopilot}. 
%
Today, many autonomous driving systems are built as modular frameworks, where each module is responsible for a particular driving function and is trained in a supervised manner. 

The decomposition into subtasks significantly simplifies the collection of supervised training signals for each component, with the design of the individual modules inspired by how humans drive. They commonly include: perception of the scene and objects of interest \cite{shi2019pointrcnn, li2022bevformer}, motion forecasting/prediction of road agents surrounding the autonomous vehicle \cite{ivanovic2019trajectron, seff2023motionlm}, mapping the lanes in which the vehicle is on \cite{liao2022maptr, luo2023augmenting}, and planning a path for the autonomous vehicle, given all of the above information \cite{casas2021mp3, hu2023planning}. Such systems have a lot of merit, proving fast, interpretable, and naturally yielding measurable performance at each step \cite{hu2023planning, zeng2019end}. 
Building upon these successes,
the modular approach has enabled the progress and \textit{democratization} of autonomous driving systems today into commercial robo-taxis in major cities and sophisticated driver assistance features enhancing everyday vehicle safety.

Indeed, the frontier has now moved toward making these systems safer and more adaptable to the diverse challenges of real-world driving environments.
Since autonomous vehicles operate in highly diverse real-world settings with a heterogeneous user base, they inevitably encounter cases absent from their training data.
This distribution shift poses a fundamental challenge: ensuring robust performance across the long-tail of rare driving situations.
One solution is to continuously collect large volumes of data and labels to expand coverage across different situations in order to continually retrain and refine the system.
While there has been an increasing number of released self-driving datasets and challenges \cite{Geiger2012CVPR, nuscenes2019, waymo_2024}, this approach may be impractical at massive scale, as the costs of data collection and model development could be prohibitive.
This raises the question: \textit{Can we generalize the autonomous driving system to diverse, real-world scenarios in a principled and cost-efficient way?}

\begin{figure}[t]
    \centering
    \includegraphics[width=\linewidth]{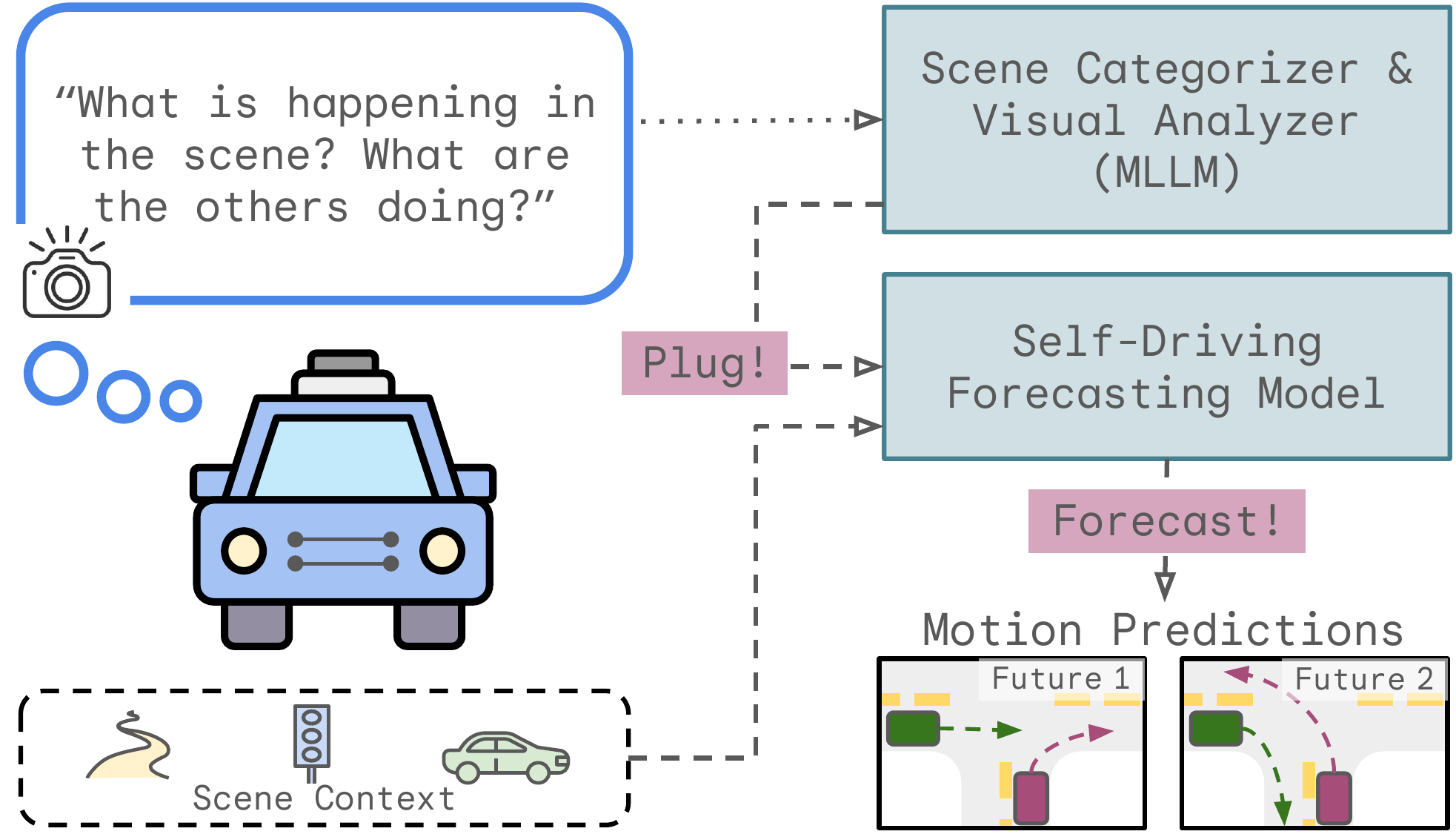}
    \caption{\textbf{Plug-and-Forecast design.} We enhance traditional autonomous driving systems by integrating MLLMs into motion forecasting models that previously relied solely on scene context from perception modules, providing comprehensive scene understanding in a zero-shot manner.}
    \label{fig:teaser}
    \vspace{-12px}

\end{figure}

One avenue of generalizability arises from the recent progress in multimodal large language models (MLLM) \cite{alayrac2022flamingovisuallanguagemodel, geminiteam2024geminifamilyhighlycapable}. MLLMs have shown strong generalist capabilities that can capture reasoning behavior, adapting to new scenarios not explicitly represented in their training data \cite{addepalli2024leveraging}. Thus, these MLLMs have the promise to augment modular driving systems with reasoning beyond their driving-specific data even in a zero-shot manner. 
In this work, we explore a method to augment an existing driving system with an MLLM, showing that information prompted from it can further improve the motion prediction driving task performance (\autoref{fig:teaser}).

Our key insight is that language is often a better descriptor for specific challenging cases and provides a powerful handle
---beyond engineering features--- 
to deal with targeted behaviors quickly. For example, edge cases such as the presence of emergency vehicles may be sparse within training data, and typical modular driving systems may produce less reliable predictions; however, a prompt can be quickly specified to explicitly target such scenarios
(\autoref{fig:gemini}).
This collection of targeted prompts can be queried into the MLLM for answers about the scene that goes \textit{beyond} the limitations of its training data, and in turn, can be distilled into learnable embeddings consumable by modular driving systems.
Specifically, we design prompts to extract information from the scene via the MLLM into a structured text format. 
This structured text is then parsed into a set of pre-defined answers, which are queried into a trained embedding space for the corresponding features.
These MLLM-extracted features are then provided as an additional input to the prediction model, augmenting it with visual reasoning information relevant to the scene.

In this work, we explore the use of both scene level information as well as road-agent specific information from these MLLMs.
We validate our approach using both types of MLLM queried features on two state-of-the-art motion behavior prediction models to demonstrate the effectiveness of such a plug-and-play approach. 
Our method, \ourslong{} (\ours{}), consistently and significantly boosts behavior prediction performance while requiring only MLLM zero-shot inference access (\ie no architecture and checkpoint access or fine-tuning needed), thereby preserving MLLMs' generalist capabilities and their ability to handle long-tailed cases \cite{zhai2024investigating}.
Specifically, our contributions are as follows:
\begin{itemize}
    \item We analyze an additional source of information, language understanding, to augment our motion prediction models.
    \item Propose a plug-and-play MLLM augmented method for the autonomous vehicle motion prediction task.
    \item Empirically demonstrate a significant and consistent improvement of performance on motion prediction task. 
\end{itemize}

\section{Related Works}
\label{sec:related}

\vspace{-10px}
\mypara{Motion forecasting.}
Among the various tasks in autonomous driving, motion forecasting/prediction is a crucial area of research, as it models the behavior of both the autonomous vehicle and other agents on the road. Early approaches to motion predictions~%
\cite{casas2018intentnet,multipath_2019,hong2019rules,biktairov2020prank}
involve rasterizing scenes into 2D images, which are then processed by convolutional neural networks (CNNs). As the field advances, research moves forward representing road elements leveraging sparsity -- such as bounding boxes, road graphs, and traffic lights -- as graph nodes, which can then be processed using graph neural networks (GNNs)%
~\cite{casas2020spagnn,liang2020learning} 
or recurrent neural networks (RNNs)%
~\cite{gupta2018social,tang2019multiple}.
Some methods adopt causal models with autoregressive trajectory prediction~\cite{rhinehart2019precog,seff2023motionlm}, allowing for sequential updates that refine predictions based on prior outcomes. In this work, we integrate MLLMs with two latest motion prediction models, Wayformer~\cite{nayakanti2023wayformer} and MotionLM~\cite{seff2023motionlm}, merging the strengths of well-established motion prediction frameworks with the broad capabilities of a knowledgeable generalist model.

\mypara{Multimodal Large Language Models in Autonomous Driving.}
The field of Multimodal Large Language Models (MLLMs)%
~\cite{alayrac2022flamingovisuallanguagemodel,openai2024gpt4technicalreport,geminiteam2024geminifamilyhighlycapable} 
has rapidly progressed, addressing the need for integrated understanding across diverse data modalities, including text, vision, and audio.
MLLMs show promising integration potential in autonomous driving systems. Several approaches enhance decision-making: DriveGPT4~\cite{xu2024drivegpt4} uses iterative Q\&A for explaining actions and predicting controls from structures data as text prompts to GPT-4~\cite{openai2024gpt4technicalreport}; 
DriveLM~\cite{sima2023drivelm} and DriveVLM~\cite{tian2024drivevlm} apply MLLMs to graph-based VQA and chain-of-thought reasoning respectively. 
Powered by Gemini~\cite{geminiteam2024geminifamilyhighlycapable}, Waymo’s EMMA model~\cite{hwang2024emmaendtoendmultimodalmodel}  directly maps raw camera sensor data into various driving-specific outputs, including planner trajectories, perception objects, and road graph elements, in a unified language space. Unlike prior work, we demonstrate that state-of-the-art motion forecasting performance can benefit from MLLMs in a zero-shot learning manner, without fine-tuning MLLMs with additional data or human labels.
\section{Problem Setup}
In this work, we focus on the task of motion prediction.
Let $\vb{s}_t$ be the states of all agents in a scenario at time $t$. Let $\vb{m} = (\vb{r}, \vb{\tau}_{1:T})$ represent the map-traffic elements, consisting of static elements (\eg lane geometries, road boundaries, crosswalks) $\vb{r}$ and dynamic elements (\eg traffic lights, temporary road works) $\vb{\tau}_t$. In addition, we assume access to passively collected ego vehicle sensor states $\xi_t$, \eg camera measurements. We wish to predict all \textit{future} agent states ${\vb{s}_{t_0+1:T}}$ for the current time $t_0$ given past agent states ${\vb{s}_{1:t_0}}$, map $\vb{m}$, and sensor measurements ${\xi_{1:t_0}}$:

\mypara{Agent States}
The state of each agent $i$ at time $t$ is represented as $s_t^i \in \mathbb{R}^d$, encoding position, velocity, heading, and agent type. The joint state $\vb{s}_t = \{s_t^1,  ..., s_t^N\}$ consists of the states of all $N$ agents in the scene. In a modular driving system, the past and current agent states are usually provided by upstream perception modules.

\mypara{Map Road Graph States}
The map-traffic representation $\vb{m}$ consists of static road elements $\vb{r}$ encoding permanent infrastructure and dynamic traffic elements $\vb{\tau}_t$ that update with traffic conditions. Road segments $\vb{r}$ is represented as a collection of road-element poly-segments. $\vb{\tau}_t$ is the traffic light states at timestep $t$.

\mypara{Sensor States}
The collected sensor measurements $\xi_t$ provide additional scene context through ego vehicle observations, particularly from camera data. These measurements are used in complement with the past agent states to extract context-dependent visual prompts for the analysis module in \ours{}.

\section{\ourslong}
\label{sec:method}
Our goal is to design a system that can leverage the reasoning and analysis capabilities of an MLLM for the modular AV stack, particularly for the task of motion prediction. Our method, \ourslong{} (\ours{}) consists of three main components: 
A language-based Visual Semantic Analyzer (\autoref{subsec:vsa}) to extract fine-grain semantic features such as agent behaviors, intentions, or visual variants. A driving Scene Categorizer (\autoref{subsec:sc}) to parse out scene-level driving characteristics such as road type or weather. And a Transformer-based architecture augmentation (\autoref{subsec:scene-embed}) to incorporate the text into a modular AV stack. Our language-based components take advantage of large, pre-trained MLLMs and do not require fine-tuning, thus making them applicable for \textit{any} modular AV task. For this work, we focus on the motion prediction task.

\begin{figure}
    \centering
    \includegraphics[width=\linewidth]{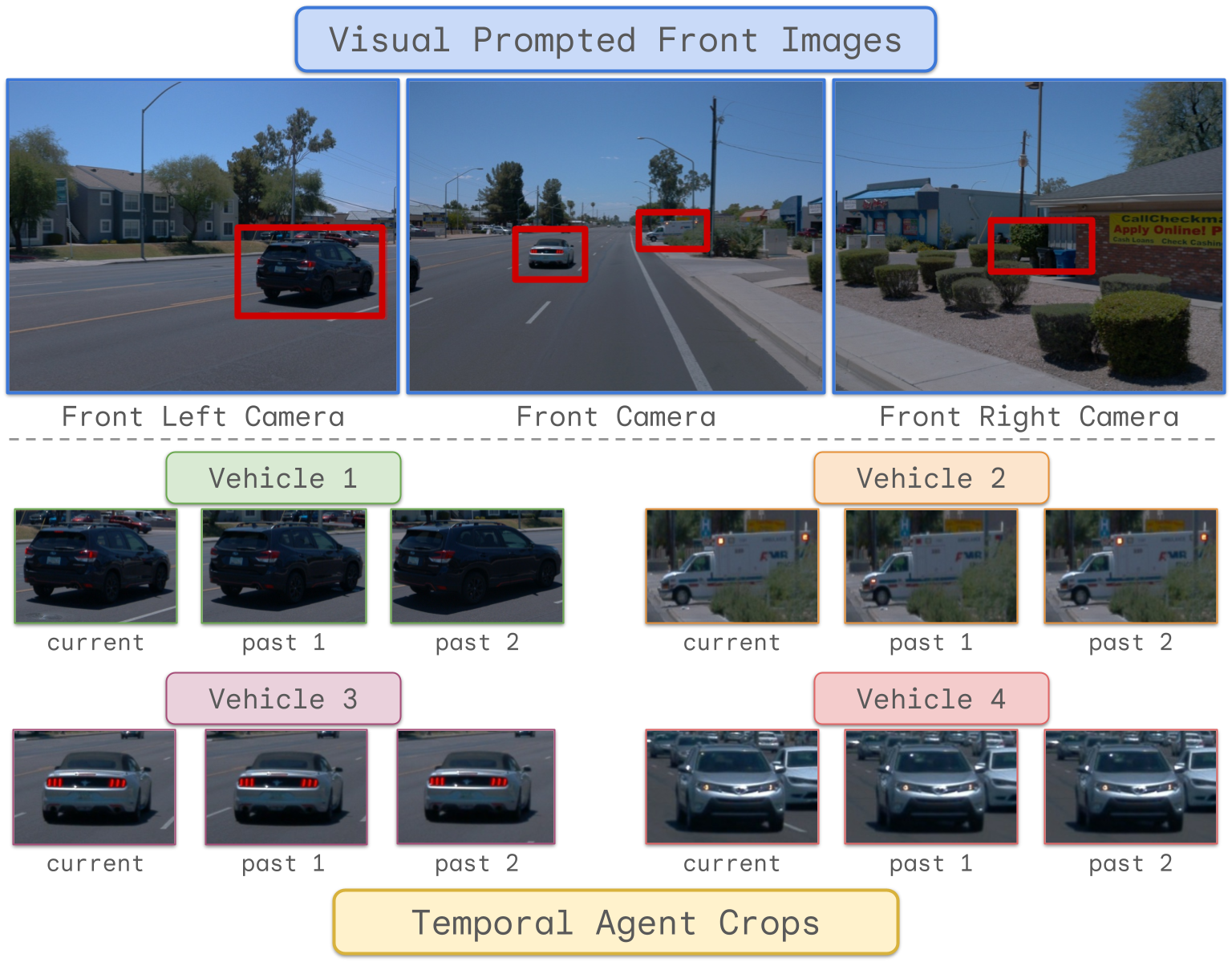}
    
    \vspace{-4px}
    \caption{\textbf{Visual Prompting Example.}}
    \label{fig:method-visual-prompting}
    
    \vspace{-16px}
\end{figure}

\begin{figure*}
    \centering
    \includegraphics[width=0.98\linewidth]{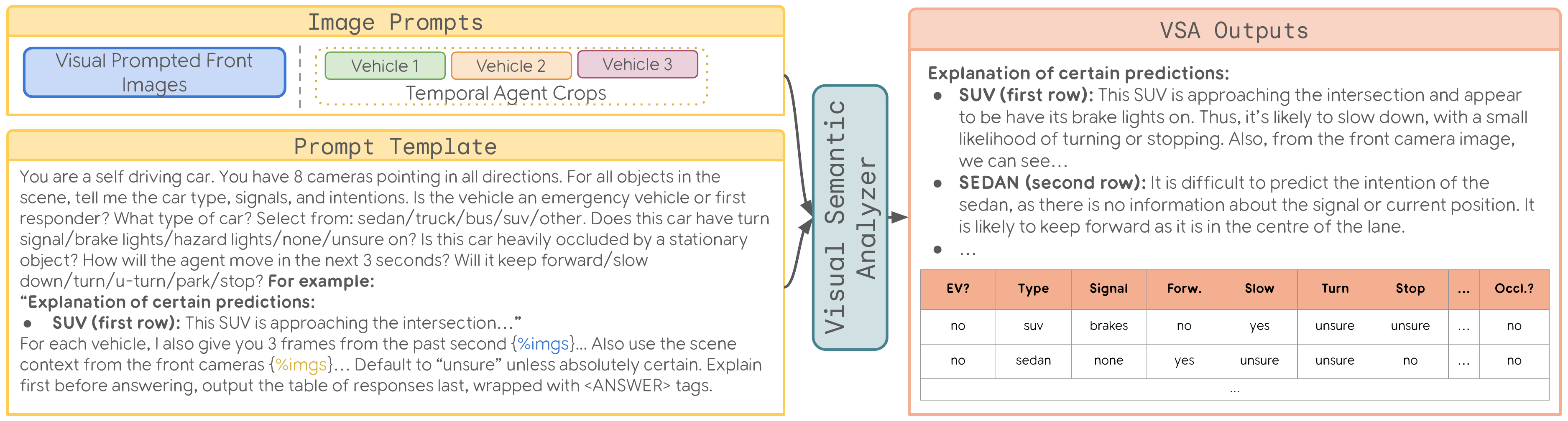}
    
    \vspace{-4px}
    \caption{
    \textbf{Example Visual Semantic Analyzer (VSA) Input and Outputs.}
    }
    \label{fig:method-vsa}
    
    \vspace{-18px}
\end{figure*}

\subsection{Visual Semantic Analyzer}
\label{subsec:vsa}
The Visual Semantic Analyzer (VSA) aims to extract agent-specific semantics useful for the downstream task of motion prediction. It takes as inputs the sensor inputs $\xi_{0:t_0}$—the rolling window of past information—and an automatically generated text prompt. It outputs a structured text representation $\vb{x}_{1:N}$ of all $N$ agents in the scene, together. Below, we describe the details of the component, and explain how $\vb{x}_{1:N}$ is produced using 
MLLMs.

\mypara{Multimodal Prompting.}
Recent progress in MLLM has extended the advances in language models into domains of images and cross-modal reasoning \cite{huang2023chatgpt, geminiteam2024geminifamilyhighlycapable}. Our work leverages the power of MLLMs in a zero-shot fashion, and does not require finetuning.
Specifically, the VSA takes a multimodal prompt, consisting of images of the current time and focused agent-following temporal crops extracted from $\xi_{0:t_0}$. We annotate the camera images from the current time using red bounding boxes around the agents-of-interest as a visual prompt. This has shown to improve their focus onto localized regions, which they otherwise may struggle to do \cite{shtedritski2023does}.
In addition, the VSA component is provided with additional visual prompts consisting of focused crops of these agents, as well as 2 past crops for additional, temporal reasoning. Because we have access to agent states from the past $\vb{s}_{1:t_0}$, we can quickly and cheaply construct a rolling memory of the camera context from the past for each agent. To reduce noisy or incomplete information getting to the MLLM, we apply filtering for object crops that are occluded or too far away. Such information is provided in addition to the text-based prompting; 
an example of the visual prompt engineering can be seen in \autoref{fig:method-visual-prompting}.

\mypara{Language Analysis of Agent Semantics.}
We wish to prompt the MLLM to obtain agent semantics useful for motion prediction. Thus, the full multimodal prompt is a mixture of both image-based visual prompting, as well as semantics specific text prompting.
To capture the variations of semantics that may be important across different agent class types (\eg between vehicle and pedestrians), we design class-specific, focused prompts. By leveraging the class type from the agent states, $\vb{s}_{t_0}$, we are able to construct a type dependent query, with the corresponding image prompts, for the scene.
In this work, we consider the vehicle and pedestrian class: For the vehicle class, we query for the presence of emergency vehicles, vehicle type, signals, and potential action intention in the next 3 seconds (keep forward, slow down, turn, stop, or parked). 
For the pedestrian class, we query for the presence of a micromobility and action intention in the next 3 seconds (jaywalk, continue on sidewalk, cross, turn, or stop/wait).
In both classes, we also query for if the agent is heavily occluded. 
To further improve the quality of the generation, we leverage generation examples and Chain-of-Thought prompting \cite{wei2022chain}.
The final output consists of per-agent chain-of-thought explanation, followed by a markdown formatted table of final answers, with an agent's feature per row. 
See \autoref{fig:method-vsa} for an example of the VSA component's input and output for the vehicle class. 
The full prompt and complete examples are provided in the supplementary.
While we experiment with a defined set of semantic tasks for MLLMs to answer, the promising capabilities of generalist models allow for easy expansion of the question set within the same proposed framework.

\begin{figure}
    \centering
    \includegraphics[width=0.98\linewidth]{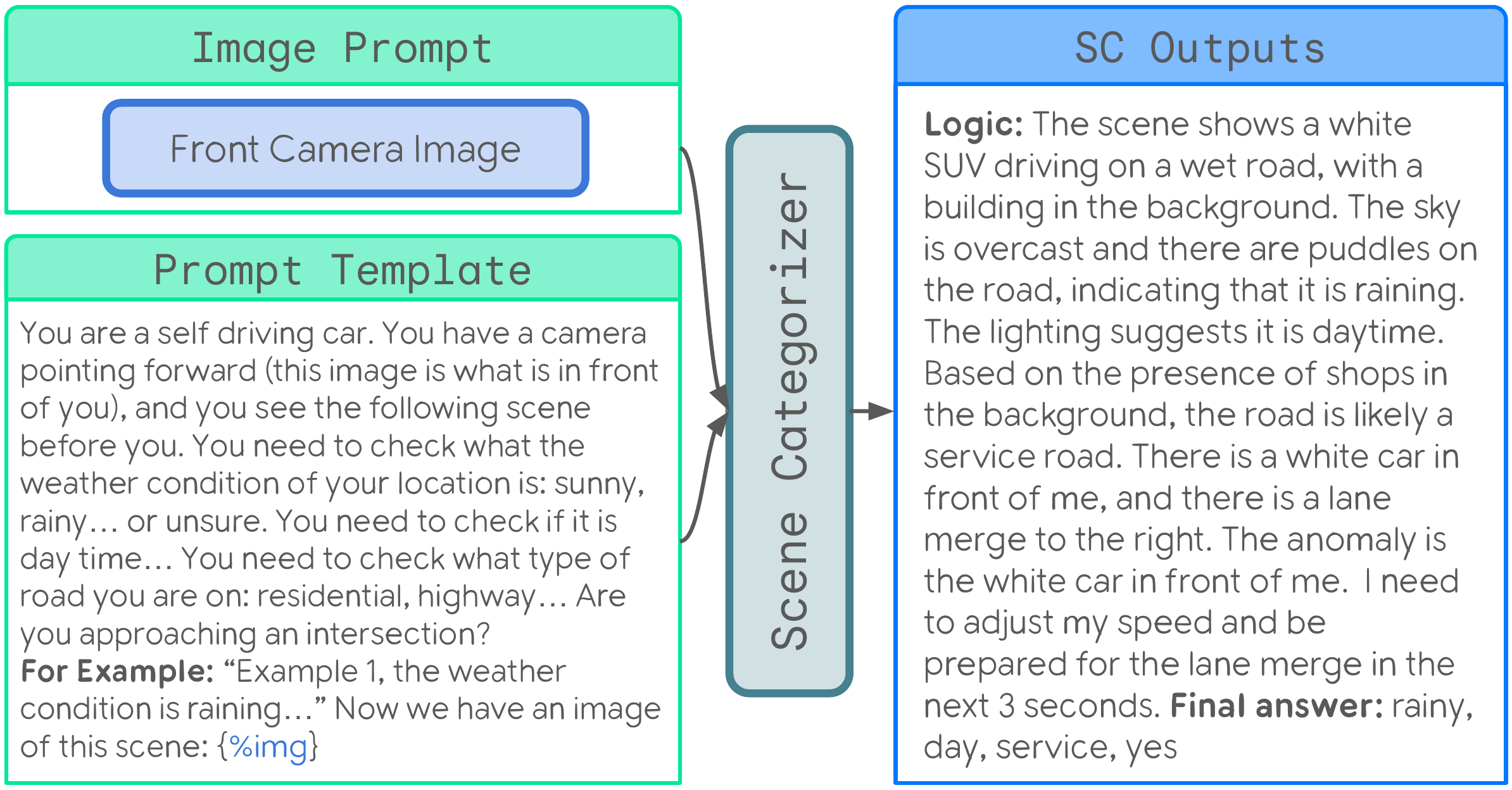}
    
    \vspace{-4px}
    \caption{\textbf{Scene Categorizer (SC) Input and Output.}}
    \label{fig:method-sc}
    
    \vspace{-8px}
\end{figure}

\mypara{Leveraging Agent-Level Structured Text.}
The generated text, while prompted to be structured for each agent and to answer specific questions, are not immediately consumable by a modular AV model. To bridge this gap, we begin by extracting the relevant row for each agent $i$, which is parsed to get an answer per question to get a list of text-based agent semantics.
We further introduce a smaller set of ``vocabulary" that are pre-specified answers to the questions; in this way, we can discretize the outputs without relying on the underlying model embeddings of the generations.
Because the generation can be incomplete or prone to minor issues, we include a default answer for all values that do not fit in the pre-defined structure. 
The final output is a multi-hot feature $\vb{x}_i$, corresponding to the answer set, for each agent $i$, that captures individual visual features and behavior intentions that is ready to be consumed by downstream models. 
There are major benefits to this design: structured outputs can be handled very easily by LLMs/MLLMs, these outputs are easy to interpret, and models can directly consume the final results as input features with minimal processing.

    
    

\begin{figure}[t]
    \centering
    \includegraphics[width=0.98\linewidth]{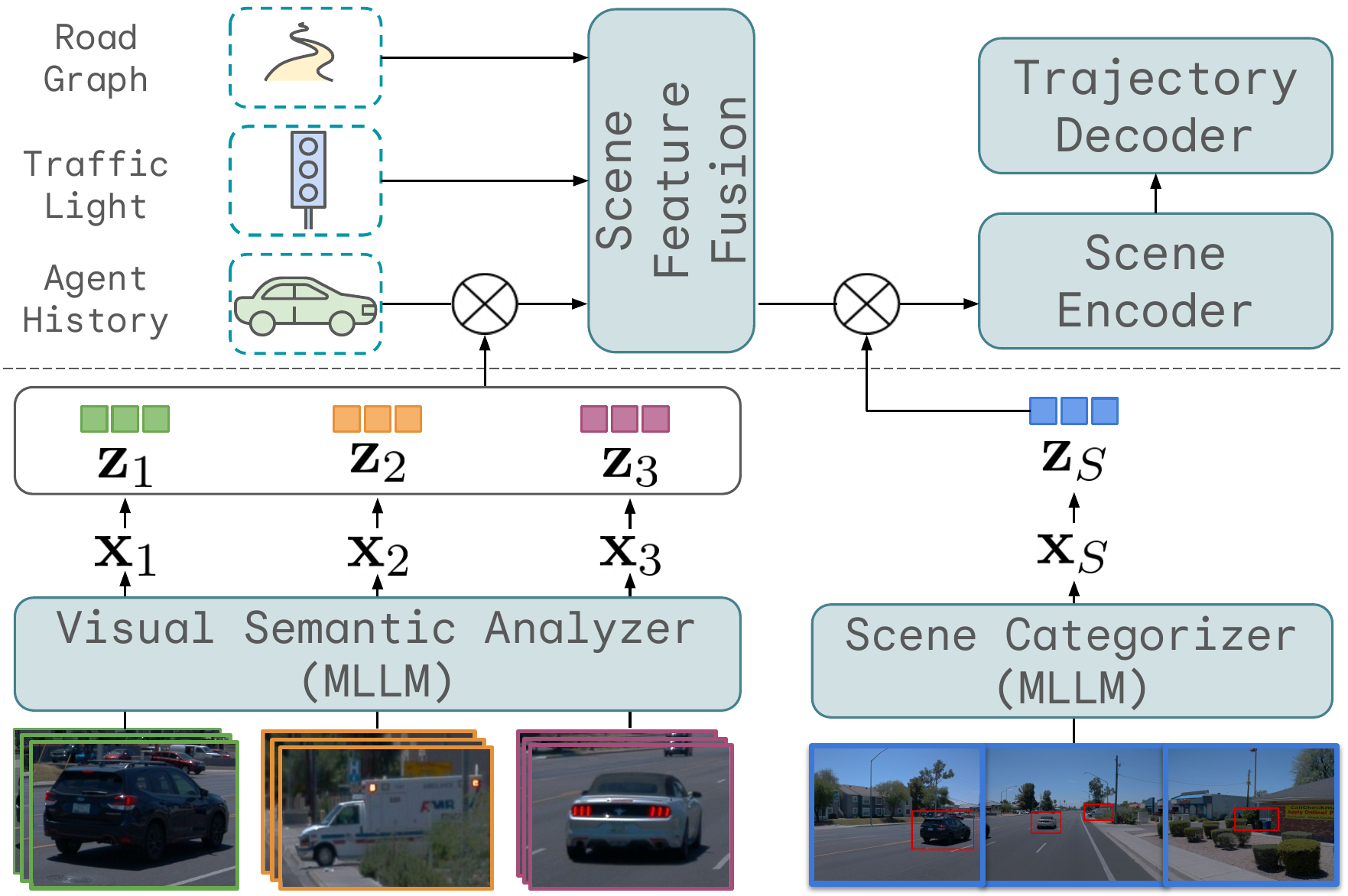}
    \caption{\textbf{Architecture of \ours{} method.} We augment a Transformer-based modular AV stack, allowing us to learn to incorporate text inputs into the stack for motion prediction. $\otimes$ stands for learned information gain operation.}
    \label{fig:method-architecture}
    \vspace{-16px}
\end{figure}

\subsection{Driving Scene Categorizer}
\label{subsec:sc}
In order to capture scene level understanding that may be present beyond individual agent-level, we add in a Scene Categorizer (SC) component that categorizes scene-level information relevant for agent behavior. 
The SC component not only provides scene-level context, but also helps with motion prediction in cases where agents are heavily occluded and may not provide good visual cues. 
Specifically, given camera inputs of the current time, $\xi_{t_0}$, it outputs another structured text representation $\vb{x}_S$, capturing holistic scene context for driving behavior.  

\mypara{MLLM for Holistic Scene Understanding.}
To extract such information, we leverage the MLLM along with $\xi_{t_0}$ and query for the current scene information important to driving, \ie weather condition of the location, time of day, road type (\eg residential or highway), and if the ego-vehicle is approaching an intersection.
We query the MLLM using a scene-level prompt, and leverage chain-of-thought reasoning to improve efficacy of the generation. The final output consists of a high-level scene reasoning, followed by the answers for each question (\autoref{fig:method-sc}).
The full SC prompt is provided in the supplementary.

\mypara{Scene-Level Structured Text Representation.}
Similar to the outputs of the VSA component, downstream modular stacks are unable to consume the text input directly. Thus, we map the final answers to the scene-level question set onto a pre-defined answer-set ``vocabulary". Similarly, we define a default ``unsure" value for capturing text that falls outside of this vocabulary. From this, we obtain for each scene a multi-hot vocabulary vector $\vb{x}_s$ that captures the scene-level categorizations necessary for driving behavior prediction.

\subsection{Plugging MLLMs into Modular AV Stacks}
\label{subsec:scene-embed}
To leverage the rich contextual information from agent-level analysis and scene categorizations from the MLLM components into the modular AV stacks, we propose an augmentation to existing Transformer-based motion prediction architectures, which are currently the predominant paradigm \cite{nayakanti2023wayformer, mu2024most,shi2022motion}. Our approach processes structured text inputs—agent-specific descriptions $\vb{x}_{1:N}$ from the VSA component and scene-level descriptions $\vb{x}_S$ from the SC component—by embedding them into a learned representation space. \ours{} then incorporates them into the Transformer prediction model via a learned information gain, applied at both the agent-feature level and to the whole scene-feature level, respectively. Finally, it uses the augmented features to predict future agent states, $\hat{\vb{s}}_{t_0+1:T}$. We visualize \ours{}'s augmented architecture in \autoref{fig:method-architecture}.


\begin{figure}[t]
    \centering
    \includegraphics[width=\linewidth]{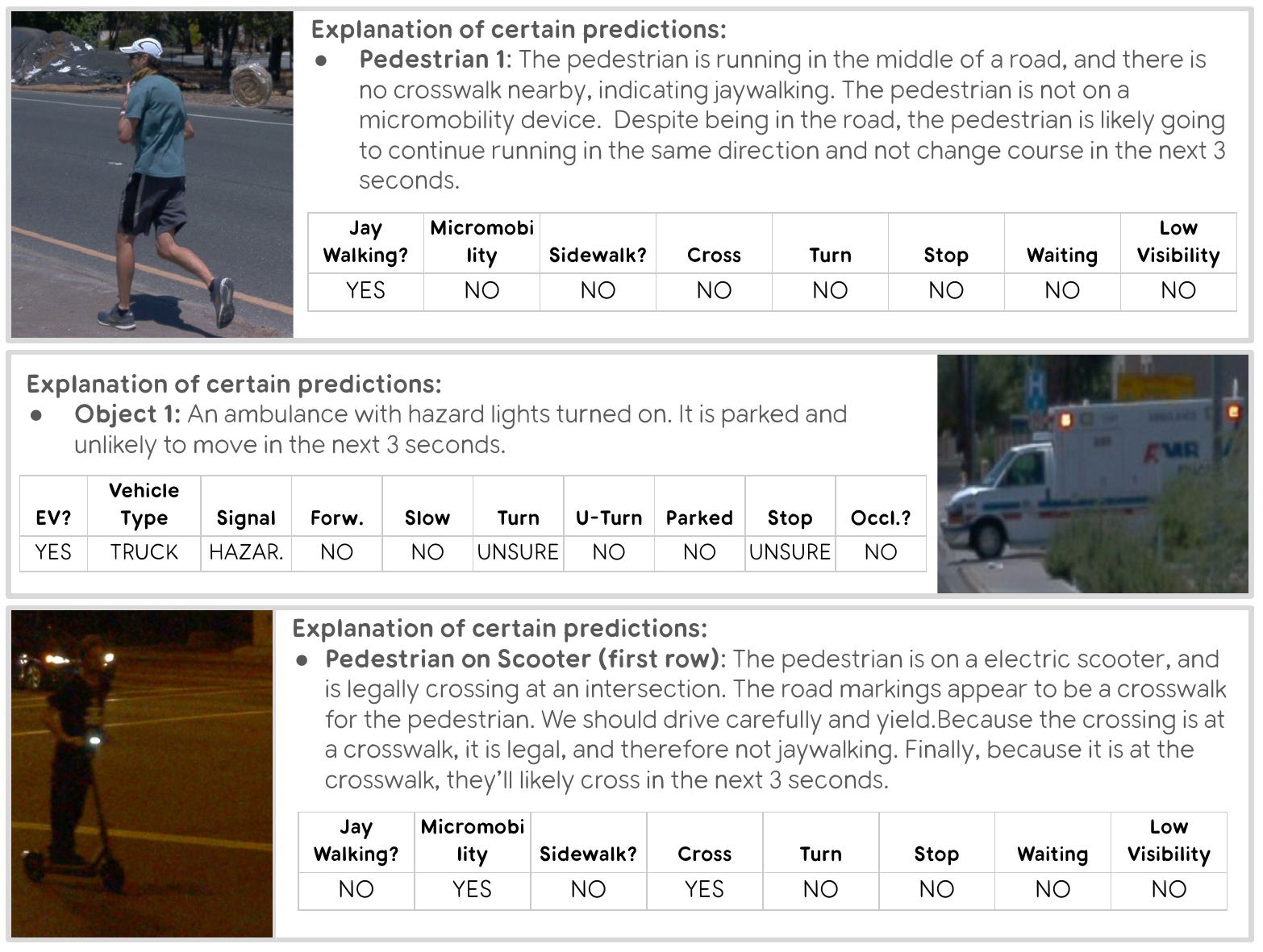}
    
    \vspace{-4px}
    \caption{\textbf{Output samples of the VSA component.} We visualize the image of the actor corresponding to the generation. Observe that the MLLM can reason about extreme cases.}
    \label{fig:gemini}
    
    \vspace{-10px}
\end{figure}

\mypara{Structured Text Embedding.}
Our \ours{} framework augments a modular AV motion prediction model via a learned incorporation of the text information from the prior components. To facilitate this, we base our modular AV model's encoder on the architecture first introduced by prior work \cite{nayakanti2023wayformer}, which has a similar Transformer base architecture that is used by many current state-of-the-art prediction models, including \cite{feng2024unitraj, seff2023motionlm}. Different from them, we learn a structured language embedding that is added onto the agent-features and subsequently scene-features, similar to how a learned positional embedding is added to improve performance. Specifically, given
$\vb{x}_{1:N}$ and $\vb{x}_S$ outputs from the VSA and SC components, we learn $\vb{z}_{1:N}$ and $\vb{z}_S$ embedding mappings into the embedding space used in the prediction model:
\begin{align}
    \forall i, ~ \vb{z}_i = \texttt{emb}_a(\vb{x}_i) \in \mathbb{R}^{d_a} \\
    \vb{z}_S = \texttt{emb}_S(\vb{x}_S) \in \mathbb{R}^{d_S},
\end{align}
where $d_a$ is the feature dimension of the agent features and $d_S$ is the dimension of the encoded scene features. We use a learned linear embedding layer $\texttt{emb}_a$ and $\texttt{emb}_S$ for the VSA and SC structured text representation, respectively. This allows the direct incorporation of the \ours{} outputs from MLLM to be used by the prediction model.

{\setlength{\tabcolsep}{8.5pt}
\begin{table*}[]

\centering
\begin{tabular}{@{}llcccccc@{}}
\toprule
Method & Reference & minADE ($\downarrow$) & minFDE ($\downarrow$) & Miss Rate ($\downarrow$) & Overlap ($\downarrow$) & mAP ($\uparrow$) & soft-mAP ($\uparrow$) \\ \midrule
MultiPath++ \cite{chai2019multipath} & ICRA 2022 & 0.978 & 2.305 & 0.440 & - & - & - \\
MTR \cite{shi2022motion} & NeurIPS 2022 & 0.605 & 1.225 & 0.137 & - & 0.416 & - \\
Wayformer \cite{nayakanti2023wayformer} & ICRA 2023 & 0.551 & 1.160 & 0.121 & - & 0.410 & 0.425 \\
MoST + Cam. \cite{mu2024most} & CVPR 2024 & 0.539 & 1.110 & 0.117 & - & 0.420 & 0.440 \\ \midrule
MotionLM$^\dagger$ \cite{seff2023motionlm} & \textit{Reproduced} & 0.574 & 1.189 & 0.139 & 0.129 & 0.382 & 0.403 \\
+ \ours{} & Ours & \textbf{0.565} & \textbf{1.166} & \textbf{0.132} & \textbf{0.129} & \textbf{0.390} & \textbf{0.413} \\ \midrule
Wayformer$^\dagger$ & \textit{Reproduced} & 0.539 & 1.111 & 0.119 & 0.128 & 0.425 & 0.446 \\
+ \ours{} & Ours & \textbf{0.528} & \textbf{1.084} & \textbf{0.113} & \textbf{0.127} & \textbf{0.437} & \textbf{0.457} \\
\bottomrule
\end{tabular}

\vspace{-4px}
\caption{\textbf{\datasetabbr{} validation set performance.} $\dagger$ marks our reproduced baselines. We add PnF on top for fair comparisons.}
\label{tab:womd-prediction}

\vspace{-4px}
\end{table*}}

\mypara{Learned Information Gain.}
In order to allow the model to selectively incorporate the sparse VSA outputs, we learn to predict an information gain ---a scalar information bottleneck clamped between $(-1, 1)$--- that is multiplied to the embedding before added into the prediction model features.
Specifically, given structured text embeddings $\vb{z}_{1:N}$ and $\vb{z}_S$, we train a small $\texttt{MLP}$ $f_{\theta_a}$ to predict the scalar gains $\alpha_{1:N}$ for each of the $N$ agents, conditioned on VSA semantics, and augment the individual agent features:
\begin{align}
    \forall i, ~ \alpha_i = \texttt{tanh}\large(f_{\theta_a}(\vb{z}_i) \large) \\
    \vb{f}_i^{'} = \vb{f}_i + \alpha_i \cdot \vb{z}_i
\end{align}
where $\vb{f}_i$ is agent $i$'s feature and $\vb{f}_i^{'}$ is the input to the encoder. These agent features are then early-fused into the other scene features via the Transformer encoder \cite{nayakanti2023wayformer}. Similarly, we train another $\texttt{MLP}$ $f_{\theta_S}$ to predict the scene-level gain $\alpha_S$ and augment the aggregated scene feature:
\begin{align}
    \alpha_S = \texttt{tanh}\large(f_{\theta_S}(\vb{z}_S) \large) \\
    \vb{f}_S^{'} = \vb{f}_S + \alpha_S \cdot \vb{z}_S
\end{align}
where $\vb{f}_s$ is the aggregated scene feature and $\vb{f}_S^{'}$ is the input to the decoder. Recall that the scene-level feature is the output of the encoder, thus contains information from the agent past states as well as map-traffic information $\vb{m}$.

This design allows the model to selectively incorporate the agent-level information since the per-agent VSA features may not be always present. Note that when feeding $0$-valued embeddings $\vb{z}$, the scalar gain $\alpha$ will be $0$, too. Additionally, this allows the model to learn a sparsification that regularizes the occasional additional features. As a bonus, the learned gain also helps to handle noisy outputs from the VSA and SG modules, thereby reducing the noise from faulty MLLM generations or hallucinations.

\mypara{Motion Prediction Training.}
To train the lightweight embedding layers, we use end-to-end training on the prediction task as supervision. From the features of the augmented prediction model encoder, the modular AV model now has an encoded representation that has context both from the traditional perception task ($\vb{s}_{1:t_0}$, $\vb{m}$) as well as context from the VSA ($\vb{x}_{1:N}$) and SC ($\vb{x}_S$) reasoning components. We leverage the query-based decoder head from \cite{nayakanti2023wayformer}, which outputs a time-series extension to a mixture-of-Gaussians,
as well as a ``valid" mask to handle variable number of predictions. 
Finally, the whole model, including the transformer-based prediction model and embedding layers for adapting MLLM outputs, is trained end-to-end on the classification and regression loss from \cite{chai2019multipath,nayakanti2023wayformer} to the poses in the ground truth observed future predictions ${\vb{s}_{t_0+1:T}}$.
From this, we construct future state per agent $\hat{\vb{s}}_{t_0+1:T}$ as our final prediction via trajectory aggregation on the output GMM following \cite{chai2019multipath, varadarajan2022multipath++}.
Note that such a design for the decoder head is simply due to the performance and ease of training, and indeed any prediction head can be swapped in without loss of generality, such as autoregressive predictors \cite{seff2023motionlm, philion2023trajeglish}. 

\section{Experimental Results}
\label{sec:experiment}


{\setlength{\tabcolsep}{5pt}
\begin{table*}[t]
\centering
\begin{tabular}{@{}lllccclccc@{}}
\toprule
\multirow{2}{*}{Method} & \multirow{2}{*}{MLLM} &  & \multicolumn{3}{c}{$K = 1$} &  & \multicolumn{3}{c}{$K = 5$} \\ \cmidrule(lr){4-6} \cmidrule(l){8-10} 
 &  &  & minADE ($\downarrow$) & minFDE ($\downarrow$) & Miss Rate ($\downarrow$) &  & minADE ($\downarrow$) & minFDE ($\downarrow$) & Miss Rate ($\downarrow$) \\ \midrule
Wayformer~\cite{nayakanti2023wayformer} & - &  & 2.471 & 6.587 & 0.830 &  & 1.137 & 2.701 & 0.542 \\ 
\multirow{2}{*}{+ \ours{}} & Qwen2-VL-7B \cite{Qwen2VL} & & 2.465 & 6.571 & 0.828 &  & 1.117 & 2.647 & \underline{0.537} \\
& Gem. 1.5 Flash \cite{geminiteam2024geminifamilyhighlycapable} &  & \textbf{2.444} & \underline{6.551} & \textbf{0.825} &  & \underline{1.102} & \underline{2.591} & 0.540 \\
 & Gem. 1.5 Pro \cite{geminiteam2024geminifamilyhighlycapable} &  & \underline{2.455} & \textbf{6.540} & \underline{0.826} &  & \textbf{1.093} & \textbf{2.574} & \textbf{0.535} \\ \midrule
MotionLM \cite{seff2023motionlm} & - &  & 2.641 & 6.960 & 0.847 &  & 1.247 & 2.972 & \underline{0.518} \\
\multirow{2}{*}{+ \ours{}} & Gem. 1.5 Flash  \cite{geminiteam2024geminifamilyhighlycapable} &  & \underline{2.598} & \underline{6.936} & \underline{0.836} &  & \underline{1.179} & \underline{2.801} & 0.526 \\
 & Gem. 1.5 Pro \cite{geminiteam2024geminifamilyhighlycapable} &  & \textbf{2.508} & \textbf{6.695} & \textbf{0.830} &  & \textbf{1.132} & \textbf{2.706} & \textbf{0.504} \\ \bottomrule
\end{tabular}

\vspace{-4px}
\caption{\textbf{NuScenes validation set performance.} We report the prediction metrics from nuScenes at $K=1$ and $K=5$.}
\label{tab:nuscenes-prediction}

\vspace{-12px}
\end{table*}
}

\mypara{Datasets.} 
\colorhighlight{We evaluate our method on two large-scale autonomous driving datasets: 
the Waymo Open Motion Dataset (WOMD) \cite{Ettinger_2021_ICCV}}
and the nuScenes Dataset \cite{nuscenes2019}. 
\colorhighlight{The \datasetfullname{}} contains over 570 hours of real-world driving scenarios across diverse locations and conditions, providing rich multi-agent trajectory data sampled at 10Hz. Each scene spans approximately 9 seconds, where we use 1 second of history and predict 8 seconds for future state prediction. The dataset includes detailed semantic map information and agent interactions across urban and suburban environments. \colorhighlight{We additionally associate the frames with their corresponding camera images to obtain inputs for the MLLM.}
The nuScenes prediction dataset comprises almost 1000 driving scenes of 20 seconds each, consisting of images and trajectory data sampled at 2Hz, collected from Boston and Singapore. We follow the standard experiment protocol of using 2 seconds of history to predict 6 seconds of future trajectories. 
Both datasets provide HD maps with semantic elements like lanes and crosswalks, while \colorhighlight{\datasetabbr{}} additionally provides traffic signals. 

\mypara{Evaluation.}
We adopt the standard metrics used in trajectory forecasting literature. For the \datasetabbr{}, we report the minimum average displacement error (minADE), minimum final displacement error (minFDE), and miss rate (MR), averaged at timesteps of 3, 5, and 8 seconds, following the Waymo Open Motion Dataset~\cite{Ettinger_2021_ICCV} leaderboard convention. On \datasetabbr{}, we additionally report mean average precision (mAP) and soft mAP, which evaluate prediction confidence scores across different motion categories (\eg, straight, turns). For the nuScenes dataset, we report the average displacement error (ADE), final displacement error (FDE), and miss rate (MR) at 2.0m threshold. Following standard practice, we report these metrics for the top $K=1$ and $K=5$ predictions, where $K$ represents the number of trajectory modes.

\mypara{Implementation Details.}
We leverage Gemini 1.5 \cite{geminiteam2024geminifamilyhighlycapable} for its relatively cheap cost, high performance, and ---most importantly--- long context window. For all experiments going forward, assume that queries to MLLM made by the VSA and SC components are to Gemini 1.5 Flash, unless otherwise specified. Accesses are made via public endpoints. We leverage the encoder-decoder framework for the modular prediction model from both \cite{nayakanti2023wayformer} and \cite{seff2023motionlm} to demonstrate the generalizability of our framework. 
{
\begin{table}[]
\setlength\tabcolsep{8pt}
\centering

\begin{tabular}{@{}lccc@{}}
\toprule
Method & minADE ($\downarrow$) & minFDE ($\downarrow$) & soft-mAP ($\uparrow$) \\ \midrule
Wayformer~\cite{nayakanti2023wayformer} & 0.819 & 1.737 & 0.305 \\
+ \ours{} & 0.753 & 1.589 & 0.330 \\ \midrule
$\Delta$ & {\color{ForestGreen}-7.99\%} & {\color{ForestGreen}-8.52\%} & {\color{ForestGreen}8.37\%} \\ \bottomrule
\end{tabular}

\vspace{-4px}
\caption{\textbf{\datasetabbr{} hardest subset performance.}}

\vspace{-4px}
\label{tab:analysis-split}
\end{table}}

{\setlength{\tabcolsep}{3.8pt}
\begin{table}[]
\centering

\begin{tabular}{@{}cccc!{\vrule width 0.2pt}ccc@{}}
\toprule
Signal & Intent & Type & Scene & minADE ($\downarrow$) & minFDE ($\downarrow$) & soft-mAP ($\uparrow$) \\ 
\midrule
\multicolumn{4}{c!{\vrule width 0.2pt}}{-} & 0.539 & 1.111 & 0.446 \\
\checkmark & \checkmark &  &  & 0.533 & 1.094 & 0.453 \\
 & \checkmark & \checkmark &  & 0.529 & 1.085 & 0.449 \\
\checkmark &  & \checkmark &  & \textbf{0.528} & 1.094 & 0.457 \\
 &  &  & \checkmark & 0.530 & 1.086 & 0.453 \\
\checkmark & \checkmark & \checkmark & \checkmark & 0.529 & \textbf{1.084} & \textbf{0.457} \\ \bottomrule
\end{tabular}

\vspace{-4px}
\caption{\textbf{Performance over different reasoning types.}}
\label{tab:mllm-ablation}

\vspace{-16px}
\end{table}
}

{
\setlength{\tabcolsep}{8pt}
\begin{table}[t]
\centering

\begin{tabular}{@{}l!{\vrule width 0.2pt}ccc@{}}
\toprule
Gain Ablation & minADE ($\downarrow$) & minFDE ($\downarrow$) & soft-mAP ($\uparrow$) \\ \midrule
None (added) & 0.531 & 1.090 & 0.454 \\
Constant & 0.531 & 1.085 & 0.449 \\
Learned (Ours) & \textbf{0.529} & \textbf{1.084} & \textbf{0.457} \\ \bottomrule
\end{tabular}

\vspace{-4px}
\caption{\textbf{Ablation on gain values used.}}
\label{tab:ablation-gain}

\vspace{-16px}
\end{table}}

\begin{figure*}[t]
    \centering
    \includegraphics[width=\linewidth]{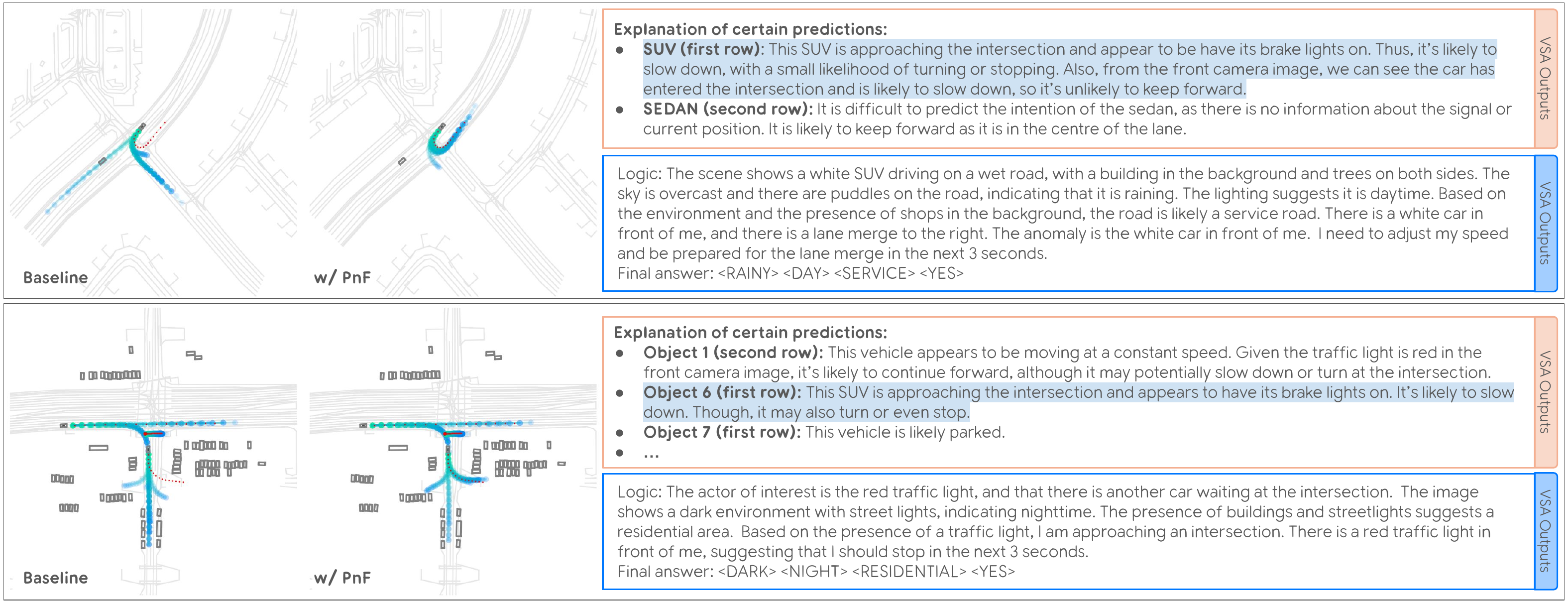}
    \caption{\textbf{Qualitative Analysis.} We visualize predictions from the Wayformer model without and with \ours{} applied. Predictions are color-coded temporally: green indicates the near future, and blue the farthest. We include outputs from the VSA and SC components corresponding to the predictions (highlighted). Our predictions better model behavior by leveraging text logic.}
    
    \vspace{-12px}
    \label{fig:qualitative}
\end{figure*}

    
    
    

\begin{figure}[h]
    \centering
    \begin{subfigure}{0.49\linewidth}
        \centering
        \includegraphics[trim={0.26cm 0 0.1cm 0}, clip, width=\linewidth]{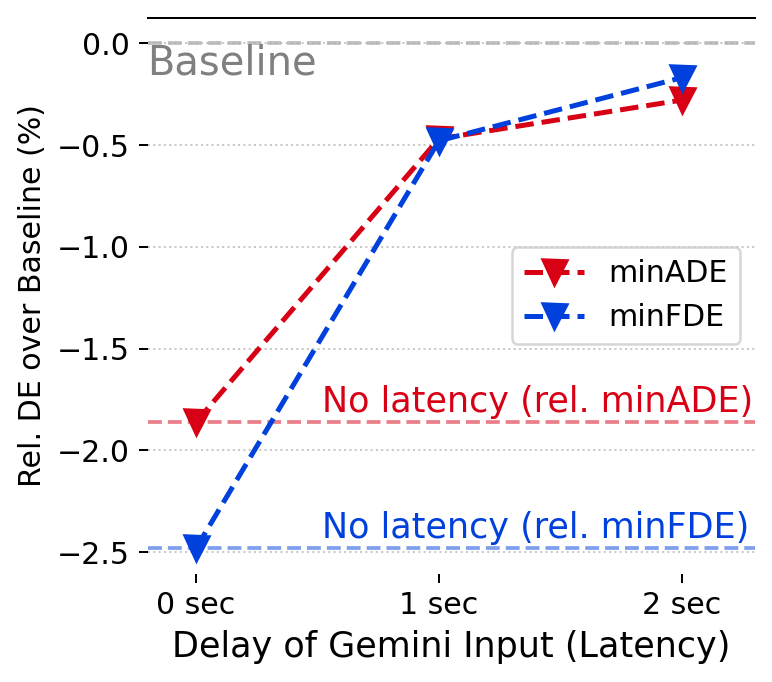}
        \label{fig:latency-de}
    \end{subfigure}\hspace{0.1em}
    \begin{subfigure}{0.49\linewidth}
        \centering
        \includegraphics[trim={0.26cm 0 0.1cm 0}, clip, width=\linewidth]{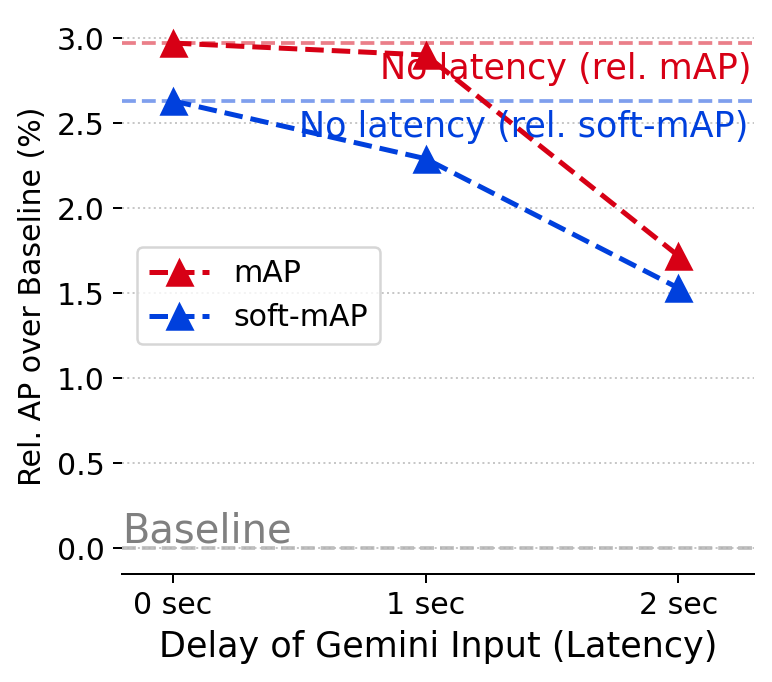}
        \label{fig:latency-ap}
    \end{subfigure}
    
    \vspace{-14px}
    
    \caption{\textbf{Method performance with latency} values at current time (0 sec), 1 sec, and 2 sec of delay on the MLLM features. We show the relative displacement errors (lower is better) on the left, and relative average precision (higher is better) relative to the baseline in {\color{gray}grey}.}
    \label{fig:latency}
    
    \vspace{-14px}
\end{figure}

\subsection{Prediction Performance}
We demonstrate the results of our method, \ours{}, as applied to the motion prediction task on the \datasetabbr{} in  \autoref{tab:womd-prediction}. 
We benchmark our work against a series of motion forecasting works.
Our method is applied to both the Wayformer model and the MotionLM model; under both settings, incorporating \ours{} consistently improves performance. Indeed, applying our method onto Wayformer beats all baselines, including MoST \cite{mu2024most}, which receives inputs from camera images but assumes accessibility of a powerful vision encoder.
Our prediction performance results on the nuScenes prediction task are reported in \autoref{tab:nuscenes-prediction}. Similarly, we observe performance gains over the baseline when \ours{} is applied. Because the nuScenes dataset is smaller, we are able to run our method using MLLM at three model sizes, Gemini 1.5 Flash and Pro \cite{geminiteam2024geminifamilyhighlycapable} as well as an open source 7-billion parameter model Qwen2-VL-7B \cite{Qwen2VL}, to demonstrate the performance across MLLM capabilities. Observe that regardless of MLLM capacity, \ours{} improves performance, particularly at larger MLLM for larger sample counts, suggesting that the VSA and SC components are able to capture different modes of motion futures.

\subsection{Method Analysis}

\mypara{Qualitative Results.}
We visualize some qualitative results of our method compared to the baseline in \autoref{fig:qualitative}. For ease of reference, we additionally provide the natural language text outputs from the VSA and the SC components corresponding to the scene prediction.
Observe that in the first example, the mode of the prediction is corrected and fixed, thanks to guidance from the analysis provided by \ours{}. 
In the second examples, the reasoning highlighted in text shows that the method is able to use the context provided to better calibrate the uncertainty between different possible action choices.

\mypara{Results on Hardest Split.}
To better understand the gains from \ours{}, we conduct an analysis on the top 10\% hardest scenarios in \autoref{tab:analysis-split}, and see that a majority of our gains (more than 3$\times$ as compared to evaluation over the whole set) is obtained there. The top 10\% hardest set is defined as the scenarios in which the baseline method achieves the highest minADE. This aligns with our intuition that our method is able to improve on the tail-end cases, thanks to the MLLMs generalizability.

\mypara{Ablations.}
To validate our design choices, we ablate our method on both the query design of the VSA and SC components in \autoref{tab:mllm-ablation}, and the architecture design choice of our information bottleneck in the input gain in \autoref{tab:ablation-gain}. Observe that with the inclusion of all the query information, the performance is the best. Removing type information reduces all metrics. Removing vehicle signal information drastically reduces the average precision for track prediction, suggesting that it is giving context into trajectory mode intentions. Including scene only information improves slightly over the baseline, but worse than including VSA features.

\mypara{Latency Analysis.} 
One key question that we aimed to answer is the effects on delayed outputs from the VSA and SC components, since currently MLLMs are unable to run in real-time due to the autoregressive nature of the generation. While there have been ample works to mitigate this, such as \cite{dao2022flashattention}, we analyze the effect of latency of structured text features on our method's performance. To simulate this for the prediction task, we run an analysis on using delayed outputs from generations for \ours{} on the relative performance improvement over the baseline in \autoref{fig:latency}. Specifically, we supply the modular AV stacks at time $t$ with outputs from the VSA and SC at time $t-k$ for $k=\{0, 1, 2\}$ seconds.
Observe that, while indeed the latency adversely affects the performance, we see positive gains even at 2 seconds of delay. Indeed, classification (\ie, mAP) still performs well with little decline in performance with up to 1 second of delay, suggesting that our method is decently robust to latency.

\section{Discussion and Future Works}
\label{sec:conclusion}
This work proposed \ours{}, a method that effectively augments autonomous driving systems through zero-shot MLLM inference and consistently improves performance across multiple state-of-the-art motion forecasting models. Our results demonstrate one of the potential uses of MLLMs to enhance modular autonomous driving systems. Several promising directions remain for future exploration. First, investigating real-time integration of MLLM outputs into driving systems could address latency concerns for deployment. Additionally, expanding the range of safety-critical scenarios and testing with a broader set of driving models would further validate the generalizability of our approach.
As we look into the future of autonomous vehicle advancements, focus should turn to leveraging information across modalities to generalize into real-world environments.






\section*{ACKNOWLEDGMENT}
\noindent
We thank the Perception Research team at Waymo for their insightful discussion and valuable feedback. Katie Luo is support in part by the AAUW Dissertation Fellowship.


\bibliographystyle{IEEEtran}
\bibliography{main}

\clearpage
\setcounter{page}{1}
\maketitlesupplementary

\newcommand{\myboxvsa}[3]{
\begin{figure*}[htbp]
    \centering
    {\normalfont
    \begin{tabular}{@{}c@{}}
        \colorbox{custombox}{%
            \begin{minipage}{\dimexpr\textwidth-2\fboxsep\relax}
                {\fontfamily{lmss}\selectfont \textbf{#1}}
            \end{minipage}%
        }\\
        \fcolorbox{custombox}{custombox!30}{%
            \begin{minipage}{\dimexpr\textwidth-2\fboxsep-2\fboxrule\relax}
                {\fontfamily{lmss}\selectfont #2}
            \end{minipage}%
        }
    \end{tabular}}%
    \caption{#1}
    \label{fig:#3}
\end{figure*}
}

\newcommand{\myboxvsaout}[2]{
    {\normalfont
    \begin{tabular}{@{}c@{}}
        \colorbox{vsaoutcolor}{%
            \begin{minipage}{\dimexpr\textwidth-2\fboxsep\relax}
                {\fontfamily{lmss}\selectfont \textbf{#1}}
            \end{minipage}%
        }\\
        \fcolorbox{vsaoutcolor}{vsaoutcolor!30}{%
            \begin{minipage}{\dimexpr\textwidth-2\fboxsep-2\fboxrule\relax}
                {\fontfamily{lmss}\selectfont #2}
            \end{minipage}%
        }
    \end{tabular}}%
}

\newcommand{\myboxsc}[3]{
\begin{figure*}[htbp]
    \centering
    {\normalfont
    \begin{tabular}{@{}c@{}}
        \colorbox{custombox2}{%
            \begin{minipage}{\dimexpr\textwidth-2\fboxsep\relax}
                {\fontfamily{lmss}\selectfont \textbf{#1}}
            \end{minipage}%
        }\\
        \fcolorbox{custombox2}{custombox2!30}{%
            \begin{minipage}{\dimexpr\textwidth-2\fboxsep-2\fboxrule\relax}
                {\fontfamily{lmss}\selectfont #2}
            \end{minipage}%
        }
    \end{tabular}}%
    \caption{#1}
    \label{fig:#3}
\end{figure*}
}

\newcommand{\myboxscout}[2]{
    {\normalfont
    \begin{tabular}{@{}c@{}}
        \colorbox{scoutcolor}{%
            \begin{minipage}{\dimexpr\textwidth-2\fboxsep\relax}
                {\fontfamily{lmss}\selectfont \textbf{#1}}
            \end{minipage}%
        }\\
        \fcolorbox{scoutcolor}{scoutcolor!30}{%
            \begin{minipage}{\dimexpr\textwidth-2\fboxsep-2\fboxrule\relax}
                {\fontfamily{lmss}\selectfont #2}
            \end{minipage}%
        }
    \end{tabular}}%
}






\section{MLLM Prompting Details}
We include the exact prompts used for the Visual Semantic Analyzer (VSA) and Scene Categorizer (SC) components in the following sections.

\subsection{VSA Prompting}
For the VSA component, we break our prompts down by vehicle class and pedestrian class, to best target challenging cases and behaviors. We present our prompt for the vehicle class in \autoref{fig:vsa-veh}, targeting difficult cases including emergency vehicles (EV), and getting additional information such as vehicle type, signaling behavior, and high-level intention approximation. Similarly, we present the prompt for the pedestrian class in \autoref{fig:vsa-ped}, where we tackle edge cases such as micromobility (scooters, skateboards, \etc), jaywalking case, and high-level intention approximation. 
In both prompts, we use chain-of-thought and examples in the prompt to encourage correct behavior. 
We include multi-modal prompting, inserting image crops and scene images, shown in Fig. 2 of the main text.

\myboxvsa{Vehicle Class Prompt (VSA)}{You are a self driving car. You have 8 cameras pointing in all directions. For all objects in the scene, tell me the car type, signals, and intentions. Is the vehicle an emergency vehicle or first responder? Select from: $<$YES$>$, $<$NO$>$, $<$UNSURE$>$. What type of car? Select from: $<$SEDAN$>$, $<$TRUCK$>$, $<$BUS$>$, $<$SUV$>$, $<$OTHER$>$. Does this car have $<$TURN SIGNAL$>$, $<$BRAKE LIGHTS$>$, $<$HAZARD LIGHTS$>$, $<$NONE$>$, or $<$UNSURE$>$ on? Is this car heavily occluded by a stationary object? Select from: $<$YES$>$, $<$NO$>$, $<$UNSURE$>$. How will the agent move in the next 3 seconds? Will it $<$KEEP FORWARD$>$, $<$SLOW DOWN$>$, $<$TURN$>$, $<$U-TURN$>$, $<$PARKED$>$, $<$STOP$>$. Select from: $<$YES$>$, $<$NO$>$, $<$UNSURE$>$.
For example:\\
**Explanation of certain predictions:**\\ \\

* **SUV (first row):** This SUV is approaching the intersection and appear to be have its brake lights on. Thus, it's likely to slow down, with a small likelihood of turning or stopping.\\
$<$ANSWER$>$ \\
$\vert$ Emergency Vehicle? $\vert$ Vehicle Type $\vert$ Signal $\vert$ Keep Forward $\vert$ Slow Down $\vert$ Turn $\vert$ U-Turn $\vert$ Parked $\vert$ Stop $\vert$ Heavy Occlusion $\vert$ \\
$\vert$---$\vert$---$\vert$---$\vert$---$\vert$---$\vert$---$\vert$---$\vert$---$\vert$---$\vert$---$\vert$ \\
$\vert$ NO $\vert$ SUV $\vert$ BRAKE LIGHTS $\vert$ NO $\vert$ YES$\vert$ UNSURE $\vert$ NO $\vert$ NO $\vert$ UNSURE $\vert$ NO $\vert$ \\
$\vert$ NO $\vert$ SEDAN $\vert$ NONE $\vert$ YES $\vert$ UNSURE $\vert$ UNSURE $\vert$ NO $\vert$ NO $\vert$ NO $\vert$ NO $\vert$ \\
$<$\textbackslash ANSWER$>$ \\ \\
For each vehicle, I also give you 3 frames from the past second:$<$img$>$ $<$img$>$ $<$img$>$ ... \\ \\
Also use the scene context: the front camera took the following image around you of the scene: Front camera:$<$img$>$ Front left camera:$<$img$>$ Front right camera:$<$img$>$ \\
Default to $<$UNSURE$>$ unless absolutely certain. Answer $<$UNSURE$>$ if you cannot see clearly. Explain first before answering, output the table of responses last, wrapped with $<$ANSWER$>$ tags.}{vsa-veh}

\myboxvsa{Pedestrian Class Prompt (VSA)}{You are a self driving car. You have 8 cameras pointing in all directions. For all pedestrians in the scene, tell me if they are jaywalking, if they have a micromobility (scooter, skateboard), and their intentions. Is the pedestrian jaywalking? Select from: $<$YES$>$, $<$NO$>$, $<$UNSURE$>$. Is this pedestrian on a micromobility vehicle, such as a scooter or skateboard? Select from: $<$YES$>$, $<$NO$>$, $<$UNSURE$>$. How will the agent move in the next 3 seconds? Will they $<$WALK SIDEWALK$>$, $<$CROSS$>$, $<$TURN$>$, $<$STOP$>$, $<$WAITING$>$. Select from: $<$YES$>$, $<$NO$>$, $<$UNSURE$>$. Is the pedestrian occluded or not visible? Select from: $<$YES$>$, $<$NO$>$, $<$UNSURE$>$.
For example:\\
**Explanation of certain predictions:**\\ \\

* **Pedestrian on Scooter (first row):** The pedestrian is on a electric scooter, and is legally crossing at an intersection. We should drive carefully and yield. Because the crossing is at a crosswalk, it is legal, and therefore not jaywalking. Finally, because it is at the crosswalk, they'll likely cross in the next 3 seconds.\\
$<$ANSWER$>$ \\
$\vert$ Jay Walking? $\vert$ Micromobility $\vert$ Walk on Sidewalk $\vert$ Cross $\vert$ Turn $\vert$ Stop $\vert$ Waiting $\vert$ Low Visibility $\vert$ \\
$\vert$---$\vert$---$\vert$---$\vert$---$\vert$---$\vert$---$\vert$---$\vert$---$\vert$ \\
$\vert$ NO $\vert$ YES $\vert$ NO $\vert$ YES $\vert$ NO $\vert$ NO $\vert$ NO $\vert$ NO $\vert$ \\
$\vert$ NO $\vert$ NO $\vert$ YES $\vert$ NO $\vert$ YES $\vert$ NO $\vert$ NO $\vert$ NO $\vert$ \\
$<$\textbackslash ANSWER$>$ \\ \\
For each pedestrian, I also give you 3 frames from the past second:$<$img$>$ $<$img$>$ $<$img$>$ ... \\ \\
Make sure to use the scene context: the front cameras took the following image around you of the scene: Front camera:$<$img$>$ Front left camera:$<$img$>$ Front right camera:$<$img$>$ \\
Use the scene context to identify where the road and sidewalks are. Default to $<$UNSURE$>$ unless absolutely certain. Answer $<$UNSURE$>$ if you cannot see clearly. Explain first before answering, output the table of responses last, wrapped with $<$ANSWER$>$ tags.}{vsa-ped}

\subsection{SC Prompting}
We provide the prompt used for the SC component in \autoref{fig:sc-scene}. We query for prediction dependent fields such as weather, time of day, and type of road. Likewise, we use chain-of-thought and examples for improved prompt performance.

\myboxsc{Scene Prompt (SC)}{You are a self driving car. You have a camera pointing forward (this image is what is in front of you), and you see the following scene before you. You need to check what the weather condition of your location is: sunny, rainy, snowy, foggy, dark, or unsure. You need to check if it is day time, evening, or night time. You need to check what type of road you are on: residential, highway, express way, service road, or other. Are you approaching an intersection? You need to check how the scene will evolve in the next 3 seconds, and what are actors of interest? What is the weather condition? Select from: $<$SUNNY$>$, $<$RAINY$>$, $<$SNOWY$>$, $<$FOGGY$>$, $<$DARK$>$, $<$UNSURE$>$. What is the time of day? Select from: $<$DAY$>$, $<$EVENING$>$, $<$NIGHT$>$. What type of road you are on? Select from: $<$RESIDENTIAL$>$, $<$HIGHWAY$>$, $<$EXPRESS$>$, $<$SERVICE$>$, $<$OTHER$>$, $<$UNSURE$>$. Are you approaching an intersection or lane merge? Select from: $<$YES$>$, $<$NO$>$, $<$UNSURE$>$.\\\\
Example 1, the weather condition is raining, and there is a tree in front of you, You should respond with:\\
Logic: There are no actors of interest, as the road is empty besides me. Based on the image, it is raining, reducing the visibility. The primary concern is the tree object that is in the road. The light is still out, which suggests it is day time. I am driving on a road that is not a road with buildings on either side, suggesting that it is a residential road. I am not approaching any intersections, or merging lanes. The anomaly is the tree branch blocking the road. This is a dangerous situation and the car needs to take action to avoid it.\\
Final answer: $<$RAINY$>$ $<$DAY$>$ $<$RESIDENTIAL$>$ $<$NO$>$\\\\
Example 2, there is an emergency vehicle in front of me, and cars surrounding me have started slowing down (brake lights are visible), You should respond with:\\
Logic: The actor of interest is the white ambulance vehicle and the grey car in front of me that is slowing down. The road ahead is straight with an ambulance with flashing lights exiting from a service road. The anomaly is the emergency vehicle that is in front of me. I see the car in front of me slowing down, suggesting an emergency situation. The image shows a clear sky and well-lit surroundings, indicating sunny weather and daytime. The presence of buildings and streetlights suggests a residential area. There is an anomaly, since there is an emergency vehicle. The ambulance, which is an emergency vehicle, is in front of me, which suggests I should be careful and yield to it in the next 3 seconds.\\
Final answer: $<$SUNNY$>$ $<$DAY$>$ $<$RESIDENTIAL$>$ $<$NO$>$\\ \\
Now we have an image of this scene:$<$img$>$}{sc-scene}

\section{Additional Qualitative Results}



\subsection{Full Ablations Tables}

\begin{table*}[t]
\centering
\begin{tabular}{@{}cccclcclcc@{}}
\toprule
\multicolumn{4}{c}{MLLM Reasoning Over}                                                                           &  & \multicolumn{2}{c}{Displ. Error}                                                      &  & \multicolumn{2}{c}{Avg. Prec.}                                                   \\ \cmidrule(r){1-4} \cmidrule(lr){6-7} \cmidrule(l){9-10} 
\multicolumn{1}{c}{Signal} & \multicolumn{1}{c}{Intention} & \multicolumn{1}{c}{Type} & \multicolumn{1}{c}{Scene} &  & \multicolumn{1}{l}{minADE ($\downarrow$)} & \multicolumn{1}{l}{minFDE ($\downarrow$)} &  & \multicolumn{1}{l}{mAP ($\uparrow$)} & \multicolumn{1}{l}{soft-mAP ($\uparrow$)} \\ \midrule
\checkmark                 & \checkmark                    & \checkmark               & \checkmark                &  & 0.529                                     & 1.084                                     &  & 0.437                                & 0.457                                     \\
\checkmark                 & \checkmark                    &                          &                           &  & 0.533                                     & 1.094                                     &  & 0.433                                & 0.453                                     \\
                           & \checkmark                    & \checkmark               &                           &  & 0.529                                     & 1.082                                     &  & 0.428                                & 0.449                                     \\
\checkmark                 &                               & \checkmark               &                           &  & 0.528                                     & 1.081                                     &  & 0.436                                & 0.457                                     \\
                           &                               &                          & \checkmark                &  & 0.530                                     & 1.086                                     &  & 0.433                                & 0.453                                     \\ \bottomrule
\end{tabular}
\caption{\textbf{Ablation table showing the different reasoning types inclusion on performance.}}
\label{tab:mllm-ablation-full}
\end{table*}
\begin{table*}[]
\centering
\begin{tabular}{@{}llcccccc@{}}
\toprule
Method &  & minADE ($\downarrow$) & minFDE ($\downarrow$) & Miss Rate ($\downarrow$) & Overlap Rate ($\downarrow$) & mAP ($\uparrow$) & soft-mAP ($\uparrow$) \\ \midrule
Wayformer \cite{nayakanti2023wayformer} &  & 0.819 & 1.737 & 0.252 & 0.138 & 0.294 & 0.305 \\
+ \ours{} &  & 0.753 & 1.589 & 0.217 & 0.137 & 0.318 & 0.330 \\ \midrule
$\Delta$ &  & {\color[HTML]{009901} -7.99\%} & {\color[HTML]{009901} -8.52\%} & {\color[HTML]{009901} -13.91\%} & {\color[HTML]{009901} -0.73\%} & {\color[HTML]{009901} 7.96\%} & {\color[HTML]{009901} 8.37\%} \\ \bottomrule
\end{tabular}
\caption{\textbf{Performance on the hardest subset on the \datasetabbr{}.}}
\label{tab:analysis-split-all}
\end{table*}
\begin{table*}[]
\centering
\begin{tabular}{@{}llcccccc@{}}
\toprule
Gain Ablation &  & minADE ($\downarrow$) & minFDE ($\downarrow$) & Miss Rate ($\downarrow$) & Overlap Rate ($\downarrow$) & mAP ($\uparrow$) & soft-mAP ($\uparrow$) \\ \midrule
None (added) &  & 0.531 & 1.090 & 0.114 & 0.127 & 0.433 & 0.454 \\
Constant &  & 0.531 & 1.085 & 0.110 & 0.127 & 0.428 & 0.449 \\
Learned (Ours) &  & 0.529 & 1.084 & 0.113 & 0.127 & 0.437 & 0.457 \\ \bottomrule
\end{tabular}
\caption{Ablation on how gain values are incorporated.}
\label{tab:ablation-gain-all}
\end{table*}

We additionally include all metrics for tables in the main text where columns may have truncated due to space. In \autoref{tab:mllm-ablation-full}, we report results corresponding to Tab. 4 of the main text, where we ablate the inclusion of different reasoning types of the VSA and SC components on final prediction performance. We include all metrics for the performance on the hardest 10\% split in \autoref{tab:analysis-split-all}. Similarly, we report all results for the ablation on how the gain controls incorporation of information from the MLLM components in \autoref{tab:ablation-gain-all}. 

\subsection{Uncertainty Estimates}
We report the standard error in \autoref{tab:womd-uncertainty}. Standard error quantifies the variability or uncertainty in a sample statistic as an estimate of the corresponding population parameter, reflecting how much the statistic is expected to fluctuate across different samples. We estimate the standard error (Std. Err.) over the entire validation set, to measure how much our method varies across different inputs. Observe that standard error is low across all metrics, and that performance gains obtained by applying \ours{} are statistically significant over the baselines. 

\begin{table*}[]
\centering
\begin{tabular}{@{}llcccccc@{}}
\toprule
\multicolumn{2}{l}{Method} & minADE ($\downarrow$) & minFDE ($\downarrow$) & Miss Rate ($\downarrow$) & Overlap Rate ($\downarrow$) & mAP ($\uparrow$) & soft-mAP ($\uparrow$) \\ \midrule
\multirow{2}{*}{Wayformer$^\dagger$} & Metric & 0.539 & 1.111 & 0.119 & 0.128 & 0.425 & 0.446 \\
 & Std. Err. & \textcolor{Dandelion}{$\pm$0.002} & \textcolor{Dandelion}{$\pm$0.004} & \textcolor{Dandelion}{$\pm$0.001} & \textcolor{Dandelion}{$\pm$0.001} & \textcolor{Dandelion}{$\pm$0.001} & \textcolor{Dandelion}{$\pm$0.001} \\
\cdashline{1-8}
\multirow{2}{*}{+ \ours{}} & Metric & 0.529 & 1.084 & 0.113 & 0.127 & 0.437 & 0.457 \\
 & Std. Err. & \textcolor{Dandelion}{$\pm$0.002} & \textcolor{Dandelion}{$\pm$0.004} & \textcolor{Dandelion}{$\pm$0.001} & \textcolor{Dandelion}{$\pm$0.001} & \textcolor{Dandelion}{$\pm$0.001} & \textcolor{Dandelion}{$\pm$0.001} \\ \midrule
\multirow{2}{*}{MotionLM$^\dagger$} & Metric & 0.574 & 1.189 & 0.139 & 0.129 & 0.382 & 0.403 \\
 & Std. Err. & \textcolor{Dandelion}{$\pm$0.004} & \textcolor{Dandelion}{$\pm$0.008} & \textcolor{Dandelion}{$\pm$0.001} & \textcolor{Dandelion}{$\pm$0.001} & \textcolor{Dandelion}{$\pm$0.001} & \textcolor{Dandelion}{$\pm$0.001} \\
\cdashline{1-8}
\multirow{2}{*}{+ \ours{}} & Metric & 0.565 & 1.166 & 0.132 & 0.129 & 0.390 & 0.413 \\
 & Std. Err. & \textcolor{Dandelion}{$\pm$0.004} & \textcolor{Dandelion}{$\pm$0.007} & \textcolor{Dandelion}{$\pm$0.001} & \textcolor{Dandelion}{$\pm$0.001} & \textcolor{Dandelion}{$\pm$0.001} & \textcolor{Dandelion}{$\pm$0.001} \\ \bottomrule
\end{tabular}
\caption{\textbf{Standard Errors on \datasetabbr.}}
\label{tab:womd-uncertainty}
\end{table*}

\section{Qualitative Results}

\subsection{Additional Qualitative Results}
\begin{figure*}[]
    \centering
    \includegraphics[width=\linewidth]{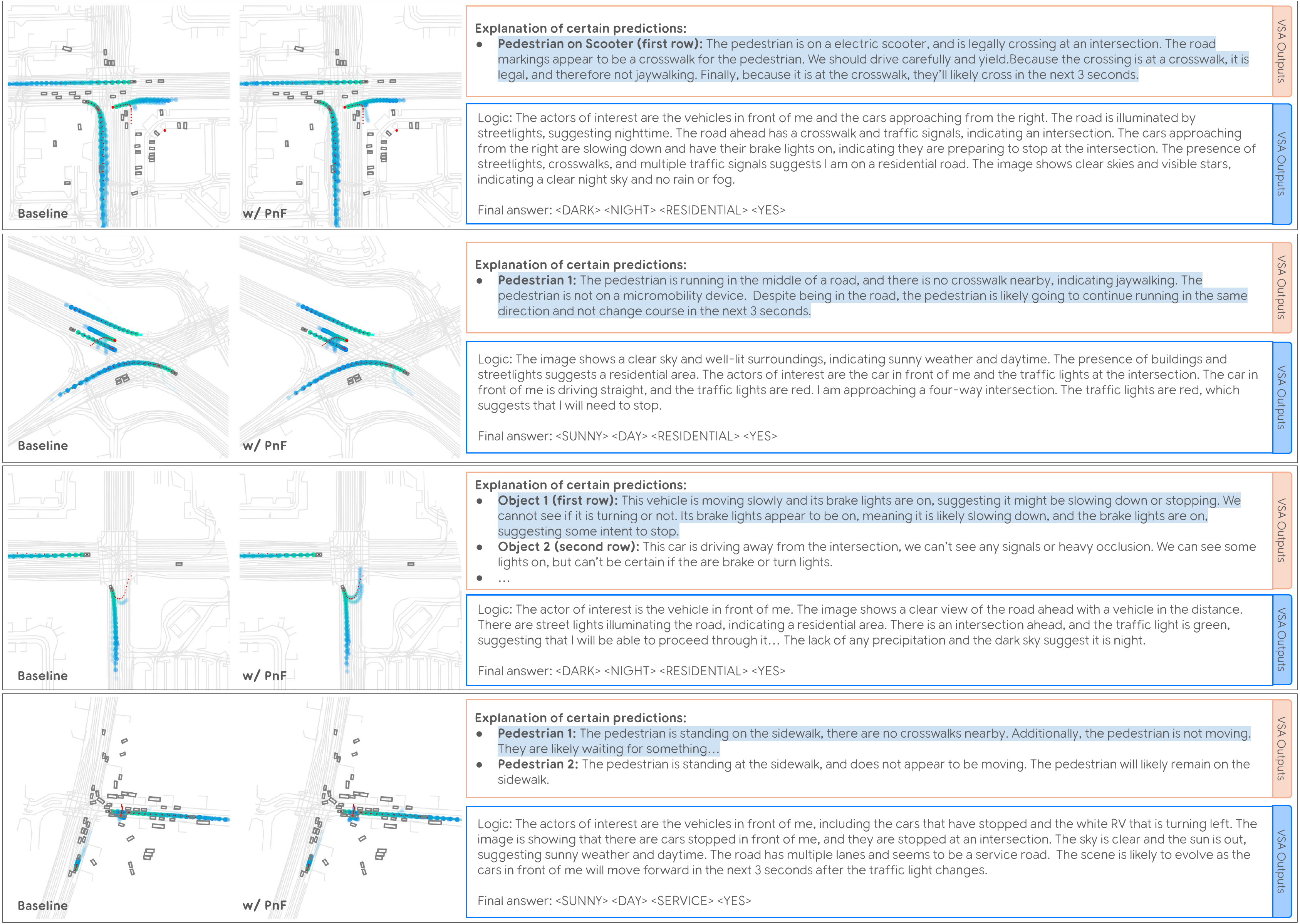}
    \caption{\textbf{Additional Qualitative Results.} We visualize additional prediction outputs on the Wayformer baseline model with and without \ours{} applied. We focus on pedestrian prediction results in this selection, where the 1st, 2nd, and 4th results demonstrate our method on pedestrian prediction. We show an additional prediction result for the vehicle class in the 3rd row.}
    \label{fig:qualitative-supp}
\end{figure*}

We visualize additional qualitative results in \autoref{fig:qualitative-supp}, with an additional focus on the pedestrian class. Observe that performance gains obtained by adding \ours{} is consistant across class, and the MLLM queried features helps to identify modes of behaviors.

\subsection{VSA and SC Outputs}
\begin{figure*}
    \centering
    
    \subfloat[Image input for visualization.]{
        \includegraphics[width=\textwidth]{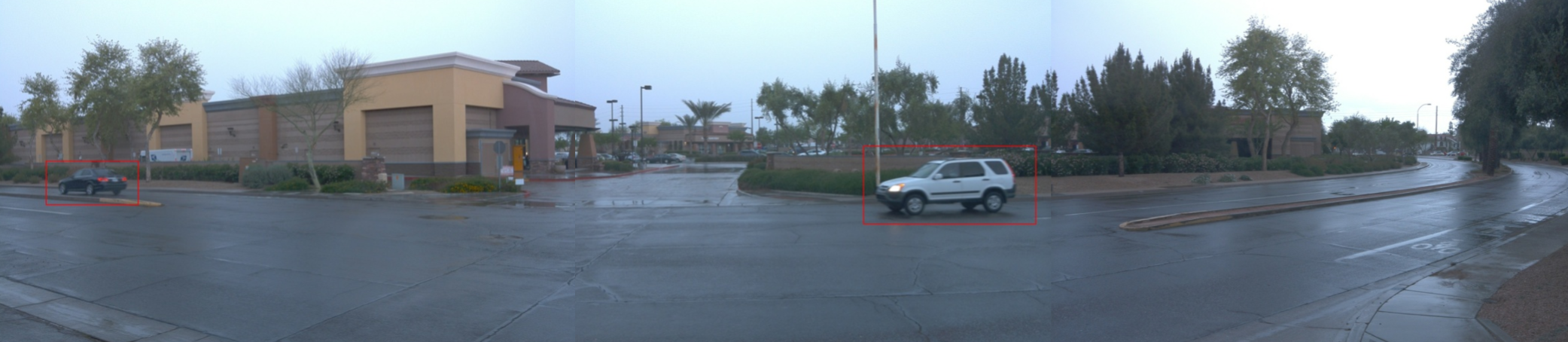}
        \label{fig:top}
    }
    
    \subfloat[VSA raw output.]{
        \myboxvsaout{VSA Output}{**Explanation of certain predictions:**\\
        * **SUV (first row):** This SUV is approaching the intersection and appear to be have its brake lights on. Thus, it's likely to slow down, with a small likelihood of turning or stopping. Also, from the front camera image, we can see the car has entered the intersection and is likely to slow down, so it's unlikely to keep forward.\\
        * **SEDAN (second row):** It is difficult to predict the intention of the sedan, as there is no information about the signal or current position. It is likely to keep forward as it is in the centre of the lane. \\ \\
        $<$ANSWER$>$\\
        $\vert$ Emergency Vehicle? $\vert$ Vehicle Type $\vert$ Signal $\vert$ Keep Forward $\vert$ Slow Down $\vert$ Turn $\vert$ U-Turn $\vert$ Parked $\vert$ Stop $\vert$ Heavy Occlusion $\vert$ \\ 
        $\vert$---$\vert$---$\vert$---$\vert$---$\vert$---$\vert$---$\vert$---$\vert$---$\vert$---$\vert$---$\vert$\\
        $\vert$ NO $\vert$ SUV $\vert$ BRAKE LIGHTS $\vert$ NO $\vert$ YES $\vert$ UNSURE $\vert$ NO $\vert$ NO $\vert$ UNSURE $\vert$ NO $\vert$\\
        $\vert$ NO $\vert$ SEDAN $\vert$ NONE $\vert$ YES $\vert$ UNSURE $\vert$ UNSURE $\vert$ NO $\vert$ NO $\vert$ NO $\vert$ NO $\vert$ \\
        $<$\textbackslash ANSWER$>$}
        \label{fig:middle}
    }
    
    \subfloat[SC raw output]{
        \myboxscout{SC Output}{Logic: The scene shows a white SUV driving on a wet road, with a building in the background and trees on both sides. The sky is overcast and there are puddles on the road, indicating that it is raining. The lighting suggests it is daytime. Based on the environment and the presence of shops in the background, the road is likely a service road. There is a white car in front of me, and there is a lane merge to the right. The anomaly is the white car in front of me.  I need to adjust my speed and be prepared for the lane merge in the next 3 seconds.\\ \\
        Final answer: $<$RAINY$>$ $<$DAY$>$ $<$SERVICE$>$ $<$YES$>$}
        \label{fig:bottom}
    }
    
    \caption{\textbf{VSA and SC component outputs demonstrated on the vehicle class.} Observe that the final output is structured in a tagged format.}
    \label{fig:mllm-output-veh}
\end{figure*}

\begin{figure*}
    \centering
    
    \subfloat[Image input for visualization.]{
        \includegraphics[width=\textwidth]{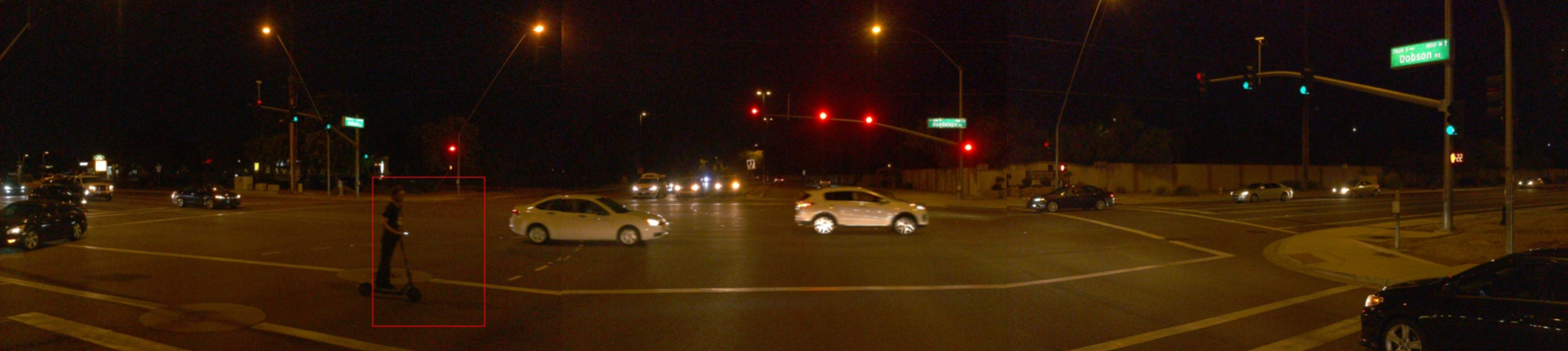}
        \label{fig:top}
    }
    
    \subfloat[VSA raw output]{
        \myboxvsaout{VSA Output}{**Explanation of certain predictions:**\\
        * **Pedestrian on Scooter (first row):** The pedestrian is on a electric scooter, and is legally crossing at an intersection. The road markings appear to be a crosswalk for the pedestrian. We should drive carefully and yield.Because the crossing is at a crosswalk, it is legal, and therefore not jaywalking. Finally, because it is at the crosswalk, they'll likely cross in the next 3 seconds. \\ \\
        $<$ANSWER$>$\\
        $\vert$ Jay Walking? $\vert$ Micromobility $\vert$ Walk on Sidewalk $\vert$ Cross $\vert$ Turn $\vert$ Stop $\vert$ Waiting $\vert$ Low Visibility $\vert$\\ 
        $\vert$---$\vert$---$\vert$---$\vert$---$\vert$---$\vert$---$\vert$---$\vert$---$\vert$\\ 
        $\vert$ NO $\vert$ YES $\vert$ NO $\vert$ YES $\vert$ NO $\vert$ NO $\vert$ NO $\vert$ NO $\vert$\\
        $<$\textbackslash ANSWER$>$}
        \label{fig:middle}
    }
    
    \subfloat[SC raw output]{
        \myboxscout{SC Output}{Logic: The actors of interest are the vehicles in front of me and the cars approaching from the right. The road is illuminated by streetlights, suggesting nighttime. The road ahead has a crosswalk and traffic signals, indicating an intersection. The cars approaching from the right are slowing down and have their brake lights on, indicating they are preparing to stop at the intersection. The presence of streetlights, crosswalks, and multiple traffic signals suggests I am on a residential road. The image shows clear skies and visible stars, indicating a clear night sky and no rain or fog.\\ \\
        Final answer: $<$DARK$>$ $<$NIGHT$>$ $<$RESIDENTIAL$>$ $<$YES$>$}
        \label{fig:bottom}
    }
    
    \caption{\textbf{VSA and SC component outputs demonstrated on the pedestrian class.} The answers are similarly structured and wrapped with tags for easy downstream use.}
    \label{fig:mllm-output-ped}
\end{figure*}

We visualize qualitative outputs of the MLLM components, VSA and SC. We include an example on the ``Vehicle" class in \autoref{fig:mllm-output-veh} and an example on the ``Pedestrian" class in \autoref{fig:mllm-output-ped}.

\section{Additional Details}
\subsection{Parameter Count Analysis}
Our method, \ours{}, consists of a learnable embedding of dimension 128 for each of the structured text outputs. Our learnable gain is a 2-layer MLP on top of the inputs. In total, this increases our parameter count from approximately 7.32 million parameters to approximately 7.33 million parameters, for an approximate 0.15\% increase on trainable parameters for the Wayformer model. We consider this a very lightweight addition, thereby increasing the practicality in adopting our method to use off-the-shelf MLLM's for improving prediction performance.

\subsection{\datasetabbr{} Dataset Details}
We provide additional details into the \datasetfullname{} (\datasetabbr). \datasetabbr{} is a real-work self-driving dataset, consisting of both camera image inputs and annotated motion tracks. The dataset is collected with 9 surround-view cameras around the vehicle. Motion trajectory annotations are labeled at 10Hz. The prediction task uses 1 second of past track information, and evaluates on 8 seconds of future information. We additionally give camera image information, similar to the NuScenes dataset \cite{nuscenes2019}. Camera information is provided up to the frame closest to the current prediction frame, at the end of 1 second past information.

\end{document}